\documentclass[runningheads]{llncs}

\usepackage{graphicx}
\usepackage{amsmath}
\usepackage{array,multirow}
\usepackage{arydshln}
\usepackage{booktabs}
\usepackage[colorlinks]{hyperref}
\begin{document}
	
	\title{Cross-View Image Retrieval - Ground to Aerial Image Retrieval through Deep Learning}
	\titlerunning{Cross-view Image Retrieval}
	%
	\author{Numan Khurshid\inst{1}\orcidID{0000-0002-8263-4781} \and
		Talha Hanif\inst{2}\orcidID{0000-0002-4067-7572} \and Mohbat Tharani\inst{3}\orcidID{0000-0001-9204-6812} \and Murtaza Taj\inst{4}\orcidID{0000-0003-2353-4462} 
	}
	\authorrunning{N. Khurshid et al.}
	%
	\institute{Computer Vision and Graphics Lab, School of Science and Engineering,\\
		Lahore University of Management Sciences, Pakistan \\ 
		\email{\{15060051\inst{1},16060073\inst{3},murtaza.taj\inst{4}\}@lums.edu.pk, l181864@lhr.nu.edu.pk\inst{2}}}
	\maketitle              
	\begin{abstract}
		Cross-modal retrieval aims to measure the content similarity between different types of data. The idea has been previously applied to visual, text, and speech data. In this paper, we present a novel cross-modal retrieval method specifically for multi-view images, called Cross-view Image Retrieval \emph{CVIR}. Our approach aims to find a feature space as well as an embedding space in which samples from street-view images are compared directly to satellite-view images (and vice-versa). For this comparison, a novel deep metric learning based solution "\emph{DeepCVIR}" has been proposed. Previous cross-view image datasets are deficient in that they (1) lack class information; (2) were originally collected for cross-view image geolocalization task with coupled images; (3) do not include any images from off-street locations.  To train, compare, and evaluate the performance of cross-view image retrieval, we present a new 6 class cross-view image dataset termed as \emph{CrossViewRet} which comprises of images including freeway, mountain, palace, river, ship, and stadium with 700 high-resolution dual-view images for each class. 
		Results show that the proposed DeepCVIR outperforms conventional matching approaches on CVIR task for the given dataset and would also serve as the baseline for future research.

		\keywords{Cross-modal Retrieval \and Cross-View Image Retrieval \and Cross-View Image Matching \and Deep Metric Learning.}
	\end{abstract}

	\section{Introduction}
	\label{sec:intro}
	
	Cross-view image matching (CVIM) attracted considerable attention of the researchers due to its growing applications in the fields of image geolocalization, GIS mapping, autonomous driving, augmented reality navigation, and robot rescue \cite{hu2018cvm,arth2011challenges}. Another key factor is the rapid increase in high resolution satellite and street-view imagery provided by platforms such as Google and Flickr. One of the most challenging task to address CVIM is to devise an effective method to fill-in the heterogeneity gap of the two types of images\cite{wang2016comprehensive,zhen2019deep}.
	
	We introduce cross-view image retrieval (CVIR) which is a special type of cross-modal retrieval, which aims to enable flexible search and collect method across dual-view images. For query image taken from one view-point (say ground-view) it searches for all the similar images taken from the other view-point (say aerial-view) in the database. The idea has evolved from the notion of cross-view image matching with one key difference. In standard cross-view image matching a ground-view image is matched to its respective aerial-view image while relying only on the content of the images. We in contrast introduce CVIR in which the system for the given query image searches for all the similar images in a database considering contextual class information embedded in visual descriptors of the images. 
	
	Common practice for conventional retrieval system is representation learning. It tries to transform images to a feature space where distance between them could be measured directly \cite{Aptoula2018DeepRSCBIR}. However, in our case these representative features must be transmuted to another common embedding space to bridge the heterogeneity gap and compute similarity between them. 
		\begin{figure}[t]
		\begin{center}
			\begin{tabular}{cccccc}
				\includegraphics[width=1.5cm]{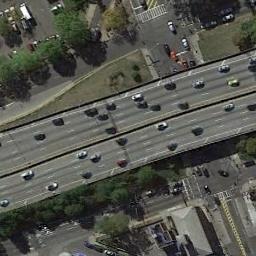}&
				\includegraphics[width=1.5cm]{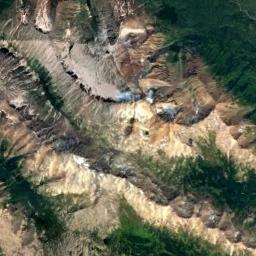}&
				\includegraphics[width=1.5cm]{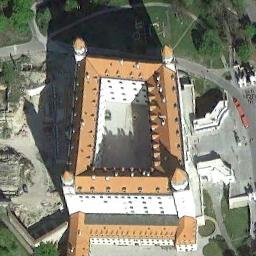}&
				\includegraphics[width=1.5cm]{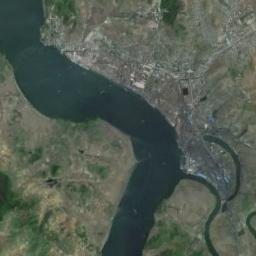}&
				\includegraphics[width=1.5cm]{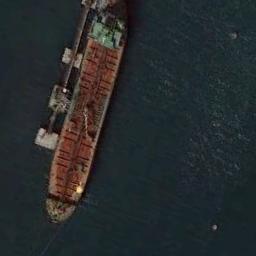}&
				\includegraphics[width=1.5cm]{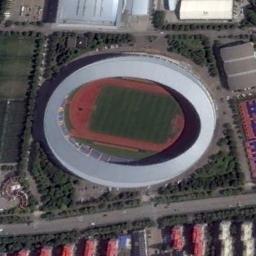}\\\\
				\includegraphics[width=1.8cm]{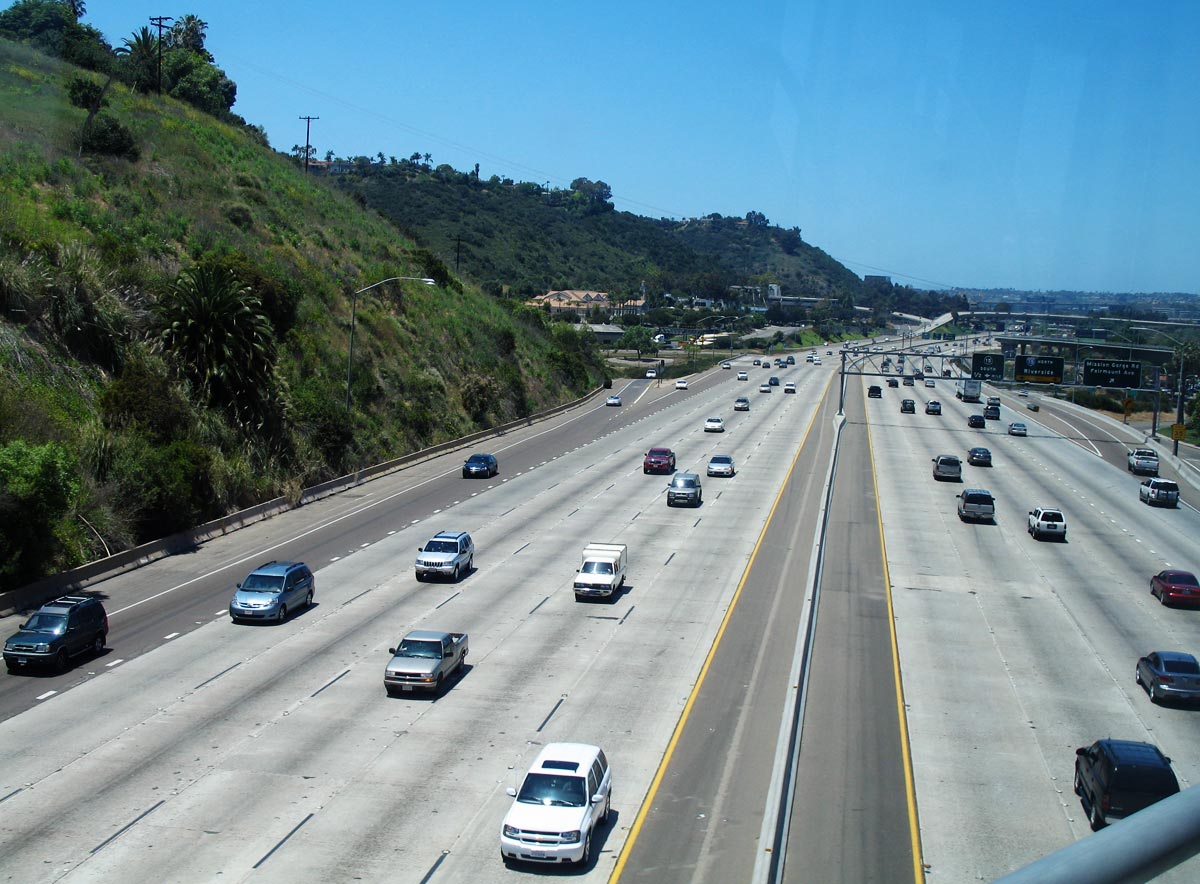}&
				\includegraphics[width=1.8cm]{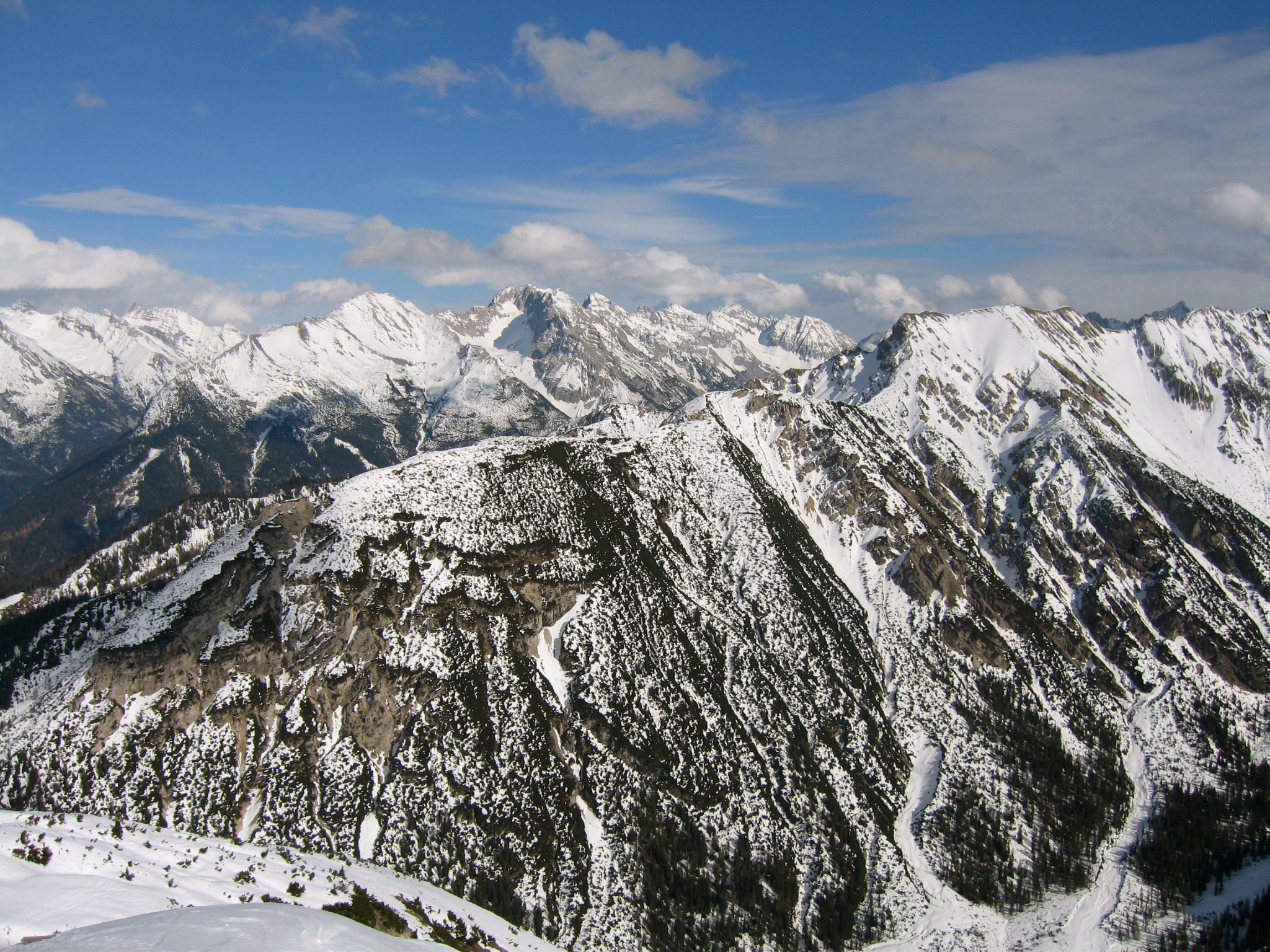}&
				\includegraphics[width=1.8cm]{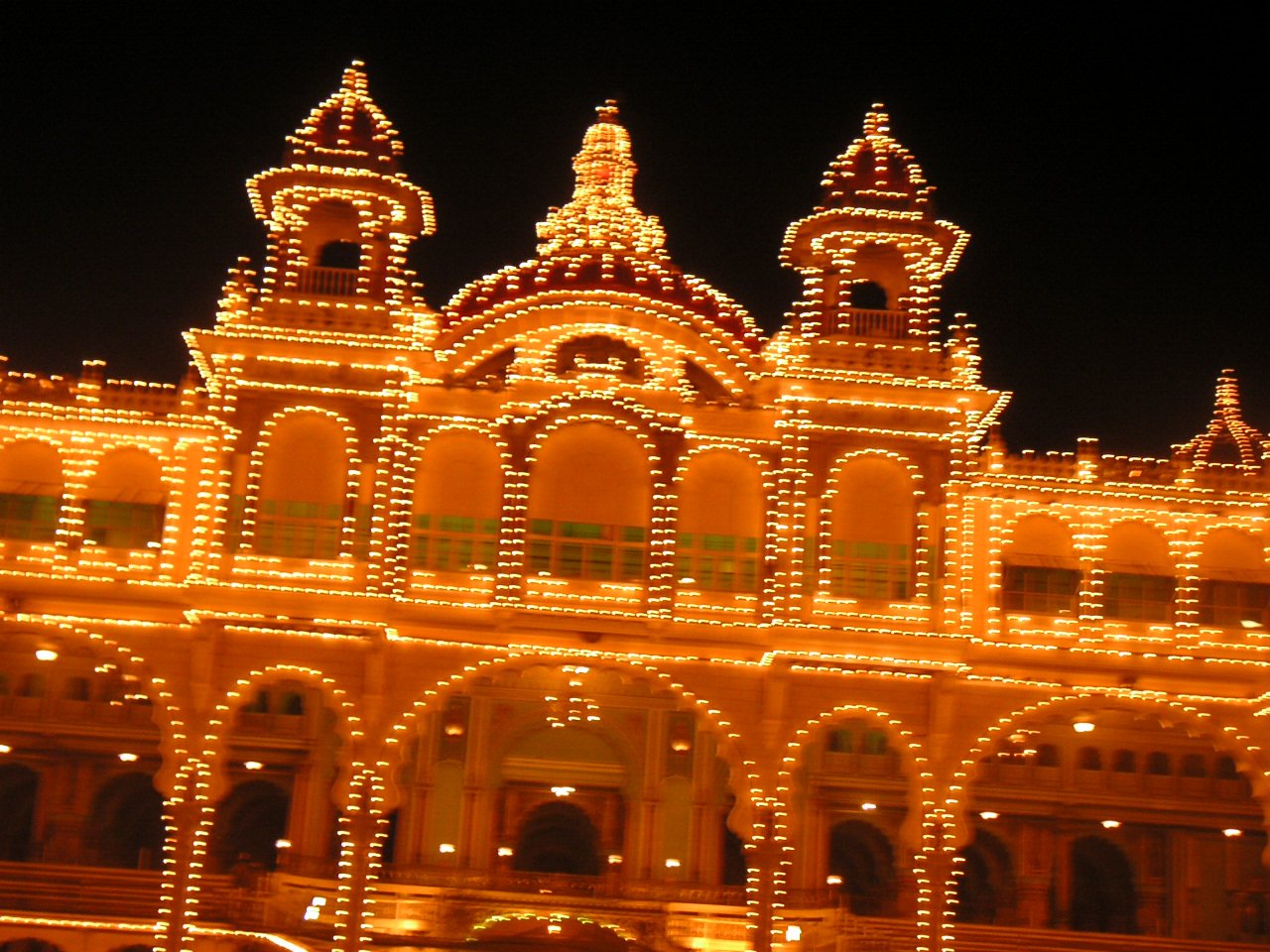}&
				\includegraphics[width=1.8cm]{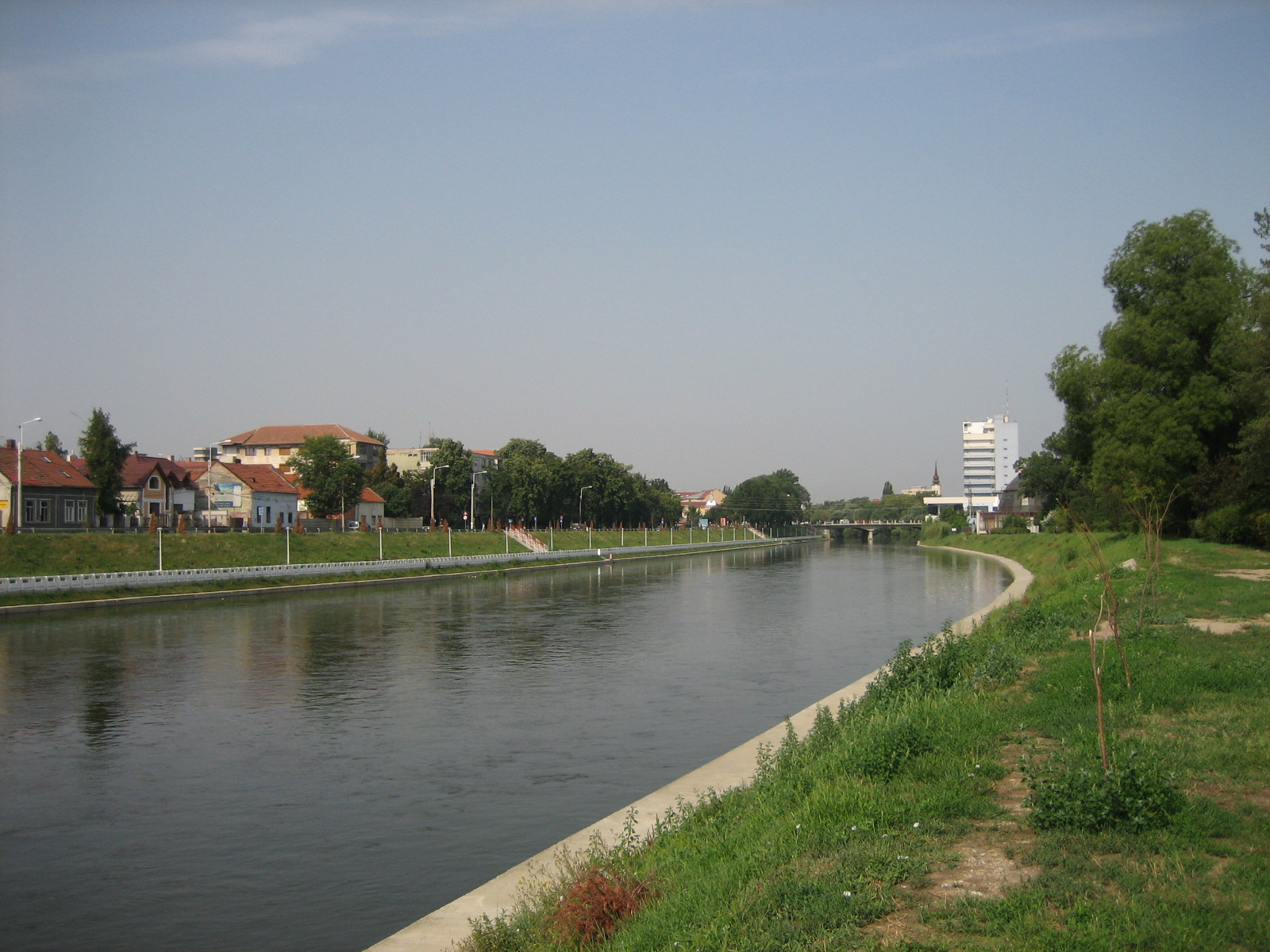}&
				\includegraphics[width=1.8cm]{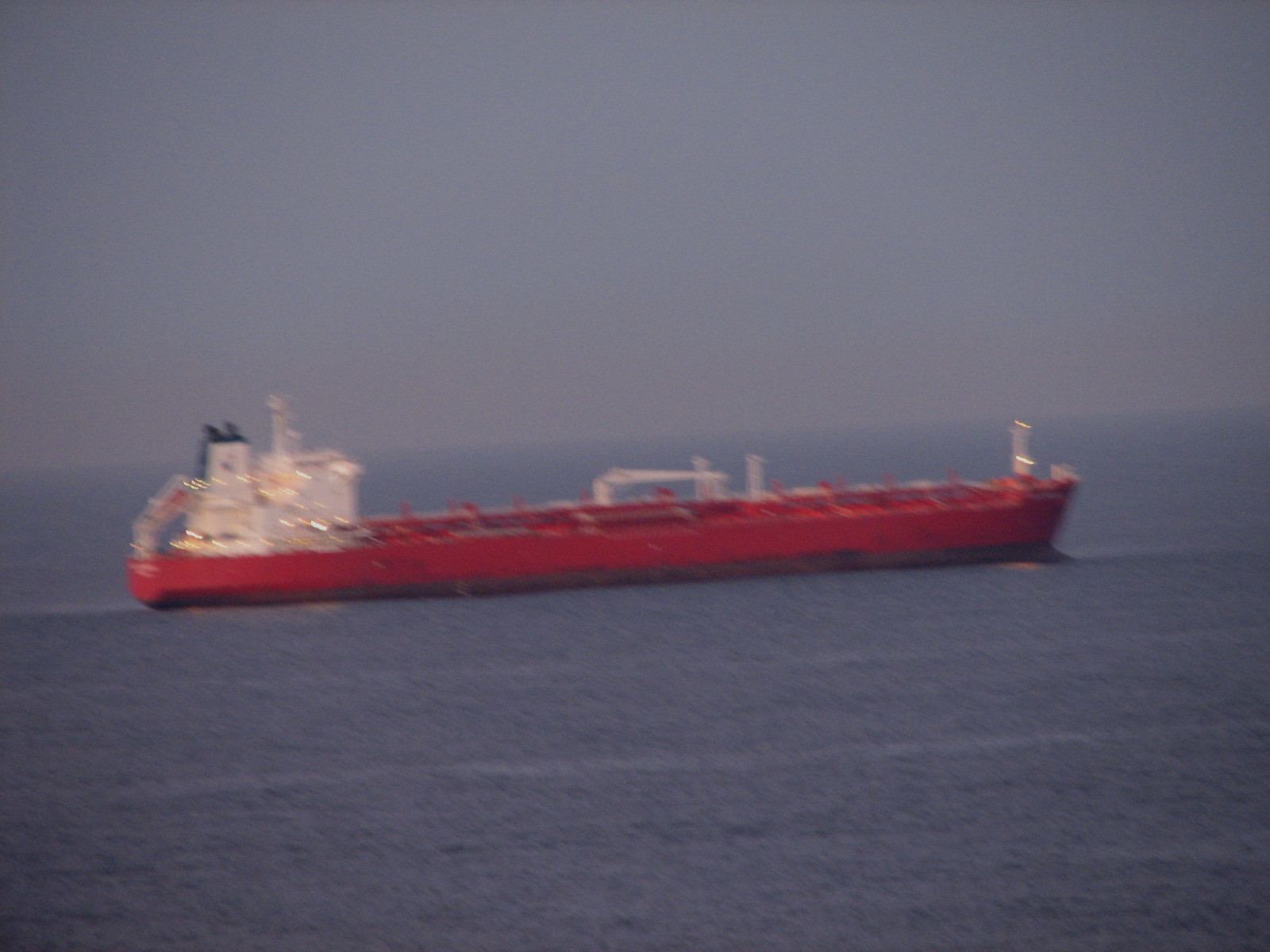}&
				\includegraphics[width=1.8cm]{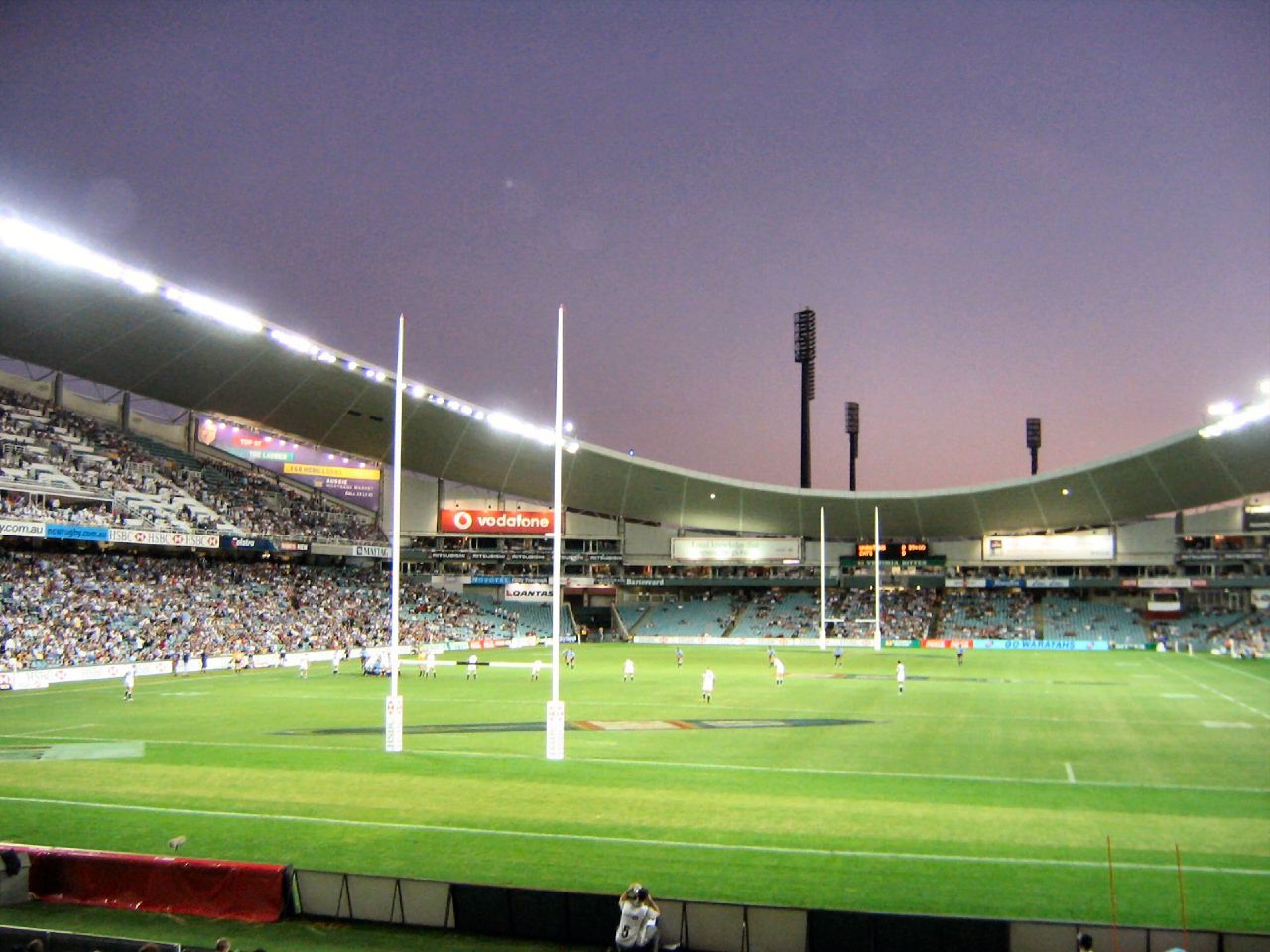}\\
				Freeway &Mountain &Palace &River &Ship &Stadium
			\end{tabular}
		\end{center}
		\caption{Some of the sample images from the developed \emph{CrossViewRet} dataset, presenting 6 distinct classes. Apart from view-point variations these images also exhibit seasonal changes, varying illumination, and different spatial resolution.}
		\label{fig:Dataset}
	\end{figure}
	
\begin{figure}[t]
		\begin{center}
			\includegraphics[width=0.95\textwidth,trim={3cm 0.35cm 2.5cm .75cm},clip]{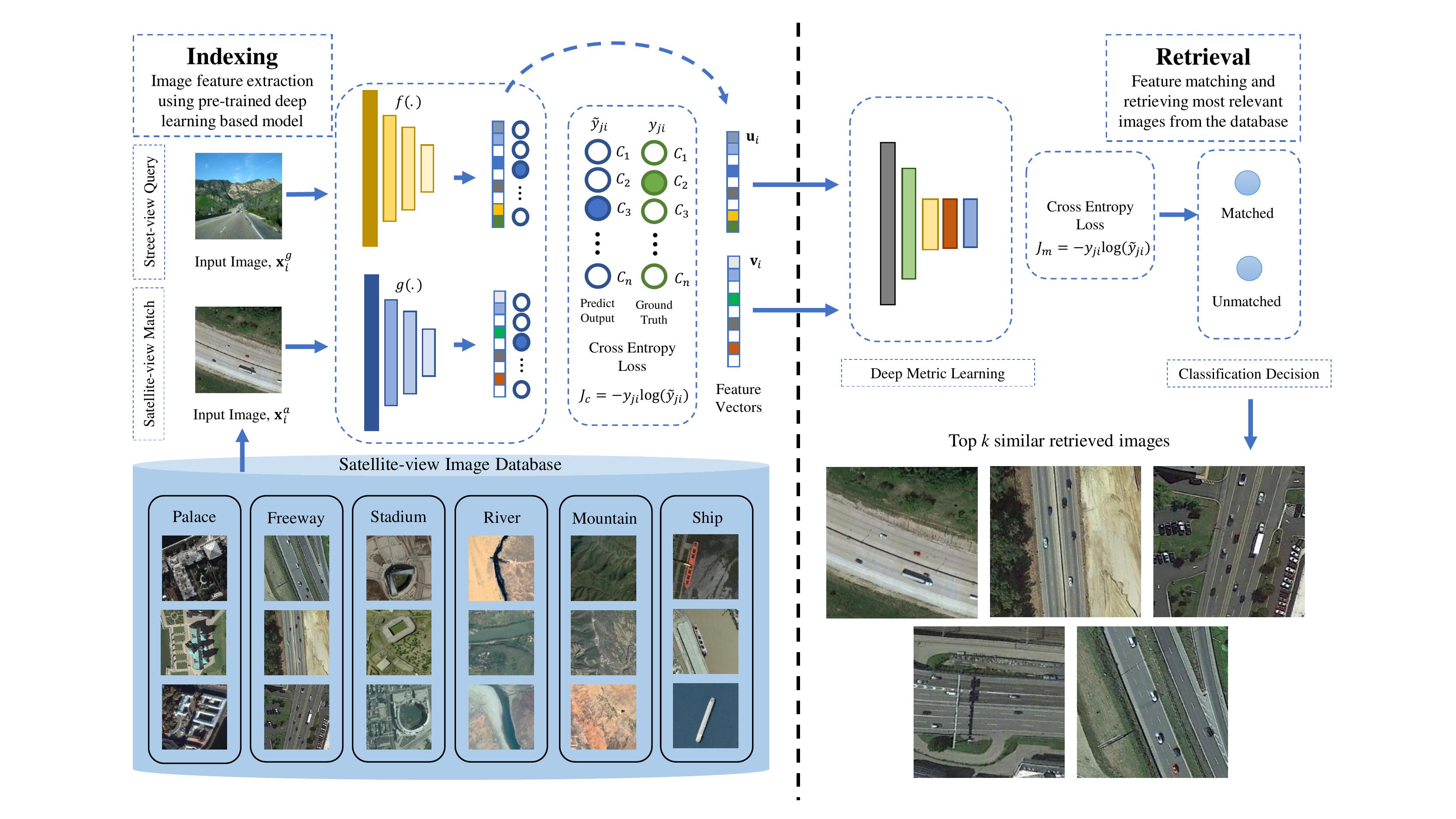}
		\end{center}
		\caption{Overall process of Cross-view Image Retrieval involving: a) Indexing step which identifies the features of the query image and image to be matched (from database), b) Retrieval step matches image features and visualize the top \textit{k} relevant images based upon the retrieval score.}
		\label{fig:CrossViewRet}
	\end{figure}

	In this paper, we present a novel cross-view image retrieval method, termed as Deep Metric Learning based cross-view image retrieval (DeepCVIR). This method aims to retain the discrimination among visual features from different semantic groups and reduces the dual-view image disparities as well. Intended to achieve this objective, class information is retained in the learned feature space and pairwise label information are retained in the embedding space for all the images. This is done by minimizing the discrimination loss of the images in both the feature space as well as embedding space to ensure the learned embeddings to be both discriminative in class information and view invariant in nature. Figure.\ref{fig:CrossViewRet} illustrates our proposed framework in detail.
	
	The remainder of this paper is organized as follows: Section \ref{literatureSurvey} reviews the related work in cross-view image matching and cross-modal learning. Section \ref{ProposedMethod} presents the proposed model including problem formulation, DeepCVIR and implementation details. Section \ref{ExperimentalSetup} explains the experimental setup including dataset while section \ref{Results} provides the results and analysis. Section \ref{Conclusion} concludes the paper. 
	
	\section{Related Work}
	\label{literatureSurvey}
	Recent applications of cross-modal retrieval especially for text, speech, and images in big-data opened new avenues which require improved solution for the recent problems. Existing technique applies cross-modal retrieval techniques to multi-modal data but do not address variety of data in any single modality such as multi-view image retrieval \cite{wei2017cross}. 
	
	Cross-view image matching could be taken as one of the potential problems for which Vo. \textit{et. al} cross-matched and geo-localized street-view images of the 11 cities of United States to their respective satellite-view images \cite{vo2016localizing}. In which experimentation using various versions of Siamese and Triplet networks for feature extraction with distance-based logistic loss have been carried out. While validating the approach on another similar dataset \emph{CV-USA} Hu. \textit{et. al} combined local features and global descriptors \cite{hu2018cvm}. One of the major short comings of both these datasets is that the street-view images are obtained from Google satellite image repository which totally ignores the off-street images. Another way to cross-match images is to detecting and matching the content of the images e.g. matching buildings in the street-view query image to the building in the aerial images \cite{tian2017cross}. This particular approach intuitively failed to perform in the area lacking any tall structures or buildings with prominent features. Researchers have even tried to predict the ground-level scene layout from their respective aerial images, however, the same approach could not be extended for accurate image matching and retrieval purpose\cite{zhai2017predicting}.
	
	Image retrieval on the other hand has already been progressively used for multi-modal matching in the field of information retrieval\cite{wang2016comprehensive}. The approach has been validated for applications to match sentences to images, ranking images, and free-hand sketch based image retrieval\cite{wei2017cross,lin2015semantics,ma2015multimodal,liu2017deep,zhen2019deep}.Moreover, metric learning networks have been previously introduced for template matching tasks \cite{xufend2015MatchNet}. We introduce cross-view image retrieval, employing the combination of metric learning and image retrieval technique for class-based cross-view image matching.
	
	\section{Proposed Method}
	\label{ProposedMethod}
	
	One of the core ideas of this paper is to identify an efficient framework for CVIR using the contextual information of the scene in image. The detailed approach is presented in four different subsections: a) Problem Formulation, b)Deep Feature Extraction, c) Feature Matching, d)DeepCVIR. Figure \ref{fig:CrossViewRet} visually explains the overall architecture of the proposed approach.

	\subsection{Problem Formulation}
	\label{CVIR}
	We focus on the formulation of CVIR problem for CrossViewRet dataset $\mathcal{D}$ without losing generality of the topic. Dataset $\mathcal{D}$ contains two subsets: ground-view images $\mathcal{D}_g$ and aerial-view images $\mathcal{D}_a$. In ground-to-aerial retrieval for the given query image $\mathrm{x}^g \in \mathcal{D}_g$ we aim to retrieve the set of all the relevant images $\mathcal{D}_{rel}$, where $\mathcal{D}_{rel} \subset \mathcal{D}_a$. Similarly, this problem could also be formulated for aerial-to-ground search and retrieval by replacing query image with $\mathrm{x}_a$ and search data as $\mathcal{D}_g$. For this purpose, we assume a collection of $n$ instances of ground-view and aerial-view image pairs, denoted as $\mathrm{\Psi} = \{{({\mathrm{x}_i}^a,{\mathrm{x}_i}^g)}\}_{i=1}^n$, where $\mathrm{x}_i^g$ is the input ground-view image sample and $\mathrm{x}_i^a$ is the input aerial-view image sample for the $i$th instance. Each pair of instances$ {({\mathrm{x}_i}^a,{\mathrm{x}_i}^g)}$ has been assigned a semantic label $\mathrm{y}_{ji}$. If $i$th instance belongs to $j$th class, $y_{ji}=0$, otherwise $y_{ji}=1$.

	\subsection{Deep Supervised Feature Learning}
	Representation learning also termed as "\emph{Indexing}" in CVIR refers to learn two functions for dual-view images containing same class information: $\mathbf{u}_i = f(\mathrm{x}_i^g;\mathrm{\phi}_g) \in \mathcal{R}^d$ for the ground-view image and $\mathbf{v}_i = f(\mathrm{x}_i^a;\mathrm{\phi}_a) \in \mathcal{R}^d $ for aerial-view image, where $d$ is the dimensionality of features in their respective feature spaces. $\phi_g$ and $\phi_a$ in the above two functions are the trainable weights of the street-view and satellite-view feature learning networks.
	Feature extraction step for the cross-view image pair is influenced by benchmark deep supervised convolutional neural networks including VGG, ResNet-50, and Tiny-Inception-ResNet-v2 pretrained networks \cite{usman2019TinyIncep}. These networks are selected due to their exceptional performance in object recognition and classification task. Unlike traditional Siamese network, here two separate feature learning networks (without weight sharing policy) are  employed for extracting features of street and satellite view images. Features acquired through this technique implicitly retain the class information of the images irrespective of their visual viewpoint. Although, these representations might not be projected in the combined feature space for both views still they share same dimensional footprint and could be compared in an embedding space through matching. Figure. \ref{fig:CrossViewRet} (left side) shows the overall indexing procedure in detail.
	
	\subsection{Feature Matching and Retrieval}
	Features of the cross-view image pair are matched either through distance computation, metric learning, or deep networks with specialized loss functions. Traditionally, matching techniques employ distance computation method of the paired data $(\mathbf{u}_i,\mathbf{v}_j)$. For instance, Euclidean distance for feature embeddings of this paired data could be computed as
	\begin{equation} \label{eqPreLiminary:distance}
	D(\mathrm{\Psi})={\|\mathbf{u}_i-\mathbf{v}_j\|}_2
	\end{equation}
	where ${\|.\|}_2$ denotes L2-norm operation.
	In distance metric learning especially contrastive embedding, a loss function implemented on top of point-wise distance operation, is minimized to learn the association of similar and dissimilar data pairs.It is mathematically computed as  
	\begin{equation} \label{eqPreLiminary:contrastiveloss}
	J_{con}=\sum_{i,j}\ell_{ij}D(\mathrm{\Psi})^2+(1-\ell_{ij}) h({\alpha-D(\mathrm{\Psi})^2})
	\end{equation}
	where $\ell_{ij} \in {0,1}$ indicates the labels of the paired data, $0$ representing similar pair and $1$ otherwise. $h(.)=max(0,h)$ is hinge loss function and $D(\mathrm{\Psi})$ is taken from (\ref{eqPreLiminary:distance}). $\alpha$ is used to penalize the dissimilar pair distances for being smaller than this predefined margin using hinge loss in the second part of (\ref{eqPreLiminary:contrastiveloss}). 
	Similarly, Mahalanobis distance between the cross-view image pair features is computed as
	
	\begin{equation}
	J_{ma}=	(  (\mathbf{u}_i - \mathbf{v}_j)' {\rm \bf C}^{-1} (\mathbf{u} - \mathbf{v}) )^{\frac{1}{2}}
	\end{equation}
	where $\vec{x}_i$ and $\vec{z}_j$ are two points from the same distribution which has covariance matrix ${\rm \bf C}$. The Mahalanobis distance is the same as the Euclidean distance if the covariance matrix becomes the identity matrix.
	variation in each component of the point.
	
	For each of the these matching measure if the retrieval score comes out to be less than the given threshold (say 0.5), the feature pair is categorized as similar and dissimilar otherwise. For image retrieval top $k$ images are visualized as relevant to the query image as shown in Figure \ref{fig:CrossViewRet}. 
	
	\subsection{DeepCVIR: A DML based framework for Feature Matching}
	\label{DML-DeepCVIR}
	The idea of transforming images from feature space to embedding space could be applied by incorporating a deep learning model technically called as deep metric learning network (DML) \cite{xufend2015MatchNet,tharani2018unsupervised}. We in this research propose a residual deep metric learning architecture optimized with the well known binary cross-entropy loss. 
	\subsubsection{Reshaping 1D features in DeepCVIR}
	To exploit the contextual information of the objects in image features we reshape 1D features $(1024\times1)$ from indexing step to 2D features $(32\times32)$ in retrieval step. 2D convolution layers are then employed to extract significant information from concatenated 2D features of the matching images. 
	\subsubsection{Residual Blocks in DeepCVIR}
	This DML network inspired from residual learning comprises the combination of two standard residual units presented in \cite{he2016identity}. The first residual unit consists of two convolution layers with an identity path while the second one comprise a $1\times1$ convolutional shortcut with two convolution layers. We tested three variations of DML for DeepCVIR. S-DeepCVIR consists of only one residual block (two residual units). For D-DeepCVIR and T-DeepCVIR, two and three stacked residual blocks are additionally used in this network, respectively. The rest of the network structure remains the same for all the three variations. Each DeepCVIR network has been terminated by the combination of three pairs of fully connected and activation layers for instigation of non-linear learning.

	\section{Experimental Setup}
	\label{ExperimentalSetup}
	Cross-view image retrieval could be inherently divided into two sub-tasks namely Steet-to-satellite retrieval and Satellite-to-street retrieval. If for the given street-view query image, satellite-view relevant images are retrieved it is referred to as Str2Sat while the vice-versa case is referred to as Sat2Str in the rest of the paper. We also investigate the effects of employing different activation functions in DML networks.
	\subsection{Dataset}
	\label{Dataset}
	In this research a new dataset \emph{CrossViewRet} has been developed to evaluate and compare the performance of DeepCVIR framework.  Previous cross-view image datasets are deficient in that they (1) lack class information about the content of the image; (2) were originally collected for cross-view image geolocalization task with coupled images; (3) were specifically acquired for the purpose of autonomous vehicles therefore they do not include any images having off-street locations. 
	CrossViewRet comprise of images containing 6 classes including freeway, mountain, palace, river, ship, and stadium with 700 high resolution dual-view images for each class. The satellite-view images are collected from the benchmark NWPU-RESISCS-45 dataset, while respective street-view images of each class are downloaded from Flickr image dataset\footnote{www.flickr.com} using Flickr API~\cite{cheng2017remote}. The downloaded street-view images are then cross checked by human annotators and images with obvious visual descriptions of the classes are selected for each class. The spatial resolution of satellite-view images is $256\times256$ and the street-view images are of variable sizes; however, they have been resized before employing for experimentation. The dataset has been made public for future use \footnote{https://cvlab.lums.edu.pk/category/projects/imageretrieval}.   
	
	CrossViewRet is a very complex dataset. Unlike existing cross-view dataset \cite{vo2016localizing,hu2018cvm} which contain ground and aerial images of the same location. We, on the other hand, do not constraint the images to be of same geo-location. Rather, we focus on visual contents in the scene regardless of any transformation in the images, weather conditions, and variation in day and night time in the scene. As shown in Fig.\ref{fig:Dataset}, the ground view in sample image of mountain class contains snow whereas its target aerial view image does not. Similarly, the ground view images of palace and stadium class are taken during night and aerial view contains day time drone images. However, the aerial view in river example is satellite image which is totally different than top view drone images. 
	
		\begin{figure}[t]
		\begin{center}
			\includegraphics[width=0.9\textwidth,trim={4cm 6cm 8cm 5cm},clip]{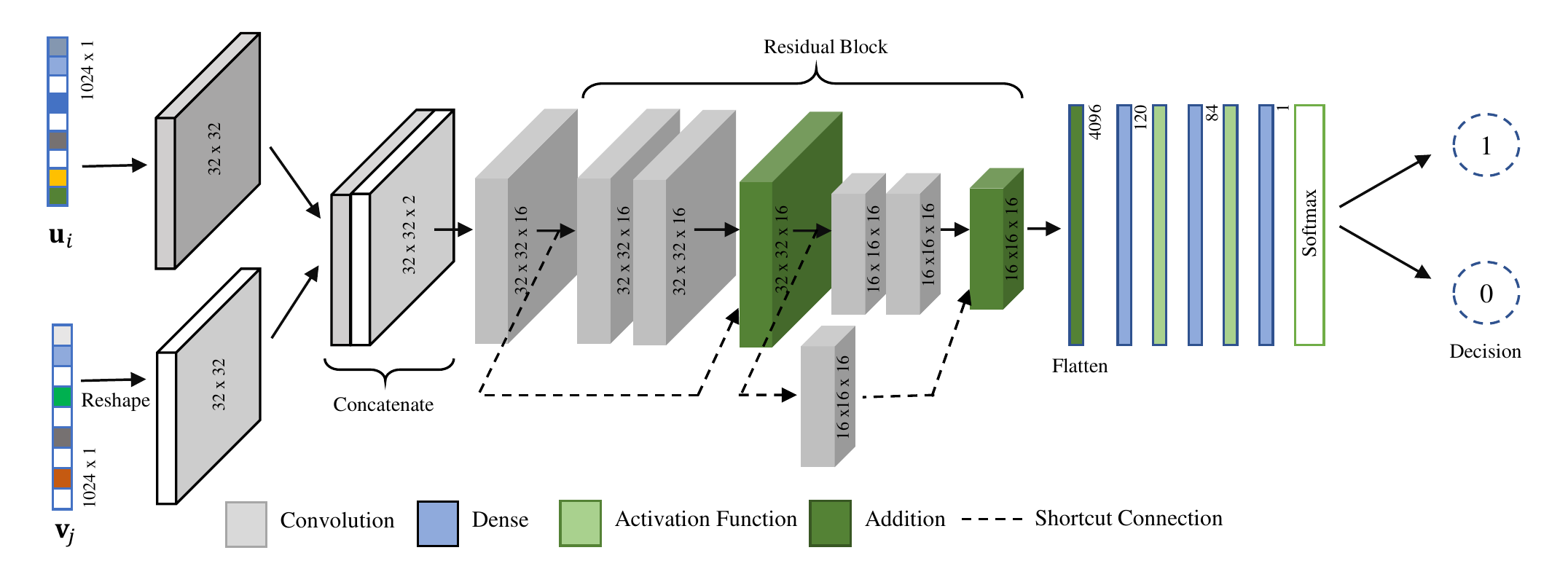}
		\end{center}
		\caption{The proposed Deep Metric Learning network (S-DeepCVIR) employed for DeepCVIR technique consists of only one residual block (two residual units). For D-DeepCVIR and T-DeepCVIR, two and three stacked residual blocks are additionally used in this network, respectively. The rest of the network structure remains the same for all the three variations. }
		\label{fig:DeepCVIR_DML}
	\end{figure}

	\subsection{Implementation Details}
	We use two independent networks for feature learning and embedding learning. In case of feature learning VGGNet, ResNet, and Inception-ResNet-v2 with pre-trained ImageNet weights are fine-tuned. Two independent sub-networks have been employed for learning the discriminating class-wise features of both the views. The architecture of the proposed DML network has been explained in section \ref{DML-DeepCVIR}. The standard 80/20 train-validation splitting criteria was used for CVIR dataset to fine-tune and train all the feature networks and variants of DML networks respectively. Query images used for evaluation were randomly taken from the validation split of the data.
	
	Deep learning networks have been trained on a Nvidia RTX 2080Ti GPU in Keras. For training feature networks, we employ Stochastic Gradient Descent (SGD) with initial learning rate 0.00001 and a learning rate decay with patience 5. For DML network, ADAM with initial learning rate of 0.001 has been used. Early stopping criteria of 15 epochs has been used to halt training for all the networks.
	
	\subsection{Evaluation Metric}
	\label{performanceMetrics}
	We evaluated the performance of cross-view image retrieval with not only the standard measures of Precision, Recall, and F1-Score but also evaluated Average Normalized Modified Retrieval Rank (ANMRR), Mean Average Precision (mAP), and P@K (read as Precision at K) \cite{Napoletano2018}. P@K is the percentage of queries which the ground truth image class are in one of the first K retrieved results. Here we only employ P@5 measure for our analysis.

	\begin{table}
		\begin{center}
			\caption{Performance comparison of features computed with state-of-the art-architectures (IncepRes-v2=Tiny-Inception-ResNet-v2).}
			\label{Table1:ComparativeFeatures}
			\resizebox{\linewidth}{!}{%
			\begin{tabular}{|l|c|c|c|c|c|c|c|} 
				\hline
				Feature Network & Similarity Measure & ANMRR$\downarrow$ & mAP$\uparrow$ & p@5$\uparrow$ & Precision & Recall & F1-Score\\ 
				\hline
				ResNet-50 &\multirow{3}{*}{\centering Euclidean}  &0.42 &0.17 &0.16 &0.50 &0.50 &0.50\\
				IncepRes-v2 &&0.42 &0.17 &0.16 &0.50 &0.50 &0.50\\
				VGG-16 &  &0.41 &0.18 &0.15 &0.48 &0.48 &0.48\\
				\cdashline{1-8}[1.0pt/2pt] 
				\addlinespace[0.1cm]
				ResNet-50 &\multirow{3}{*}{\centering Contrastive} &0.05 &0.90 &0.88 &0.50 &0.50 &0.50\\
				IncepRes-v2 & &0.40 &0.20 &0.16 &0.50 &0.50 &0.50\\
				VGG-16 &  &0.29 &0.41 &0.40 &0.50 &0.50 &0.50\\
				\cdashline{1-8}[1.0pt/2pt] 
				\addlinespace[0.1cm]
				ResNet-50 & \multirow{3}{*}{\centering Mahalanobis } &0.42 &0.17 &0.16 &0.50 &0.50 &0.50\\
				IncepRes-v2 & &0.42 &0.16 &0.15 &0.50 &0.50 &0.50\\
				VGG-16 &  &0.42 &0.17 &0.19 &0.50 &0.50 &0.50\\
				\cdashline{1-8}[1.0pt/2pt] 
				\addlinespace[0.1cm]
				ResNet-50 & \multirow{3}{*}{\centering DeepCVIR-DML } &0.03 &0.93 &0.94 &0.94 &0.94 &0.94\\
				IncepRes-v2 & &0.29 &0.22 &0.23 &0.52 &0.52 &0.52\\
			VGG-16 & &\textbf{0.02} &\textbf{0.96} &\textbf{0.97} &\textbf{0.96} &\textbf{0.96} &\textbf{0.96}\\
				\hline
			\end{tabular}}
		\end{center}
	\end{table}
	
	\section{Results}
	\label{Results}
	Validation of the proposed DeepCVIR approach for this type of challenging dataset demands extensive assessment. We therefore provide a comparative analysis of the approach using various state-of-the-art techniques as well as variants of the proposed method.
	\subsection{Deep Features and their Matching Techniques}
\begin{figure}[t]
		\begin{center}
			\begin{tabular}{cc}
				\includegraphics[width=6cm]{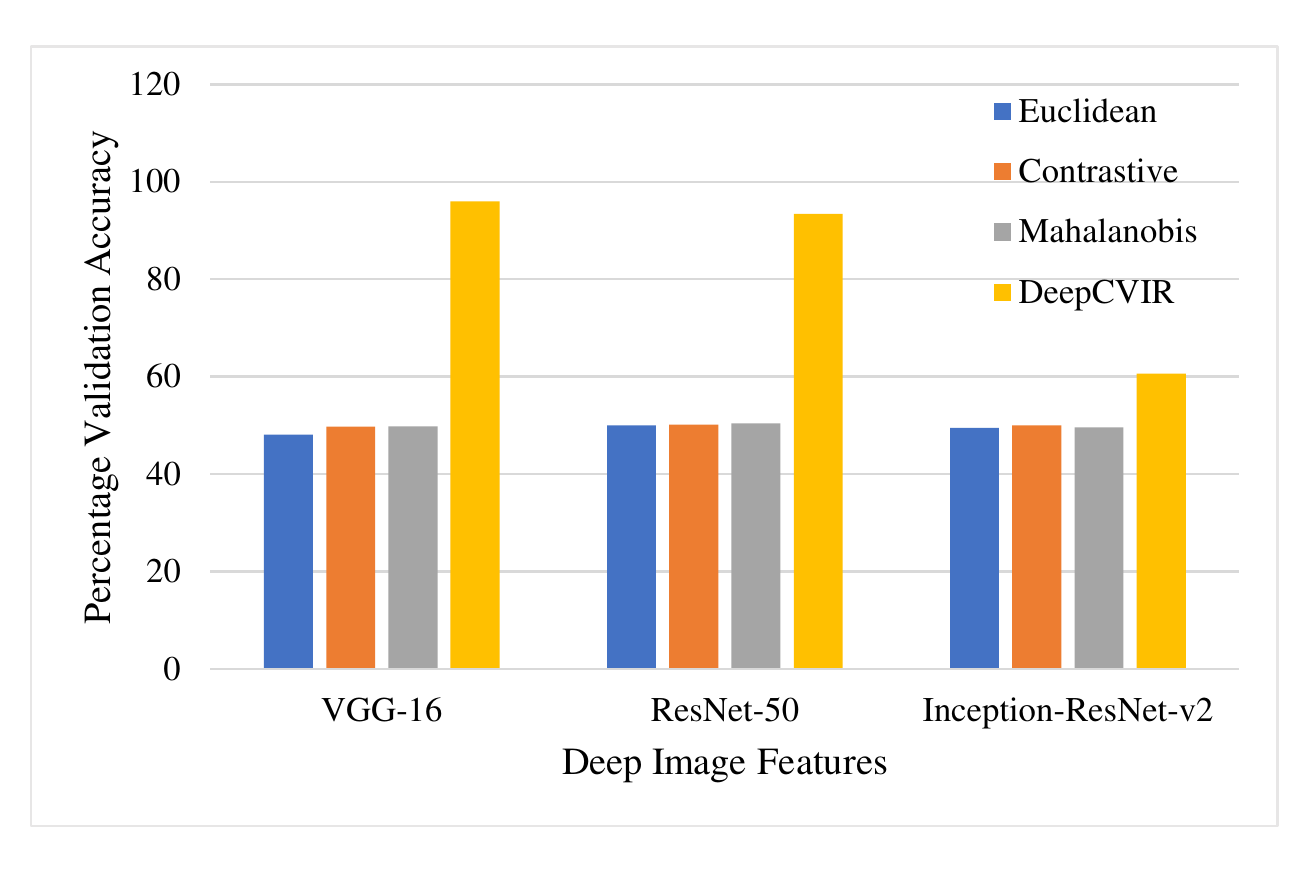}&
				\includegraphics[width=6cm]{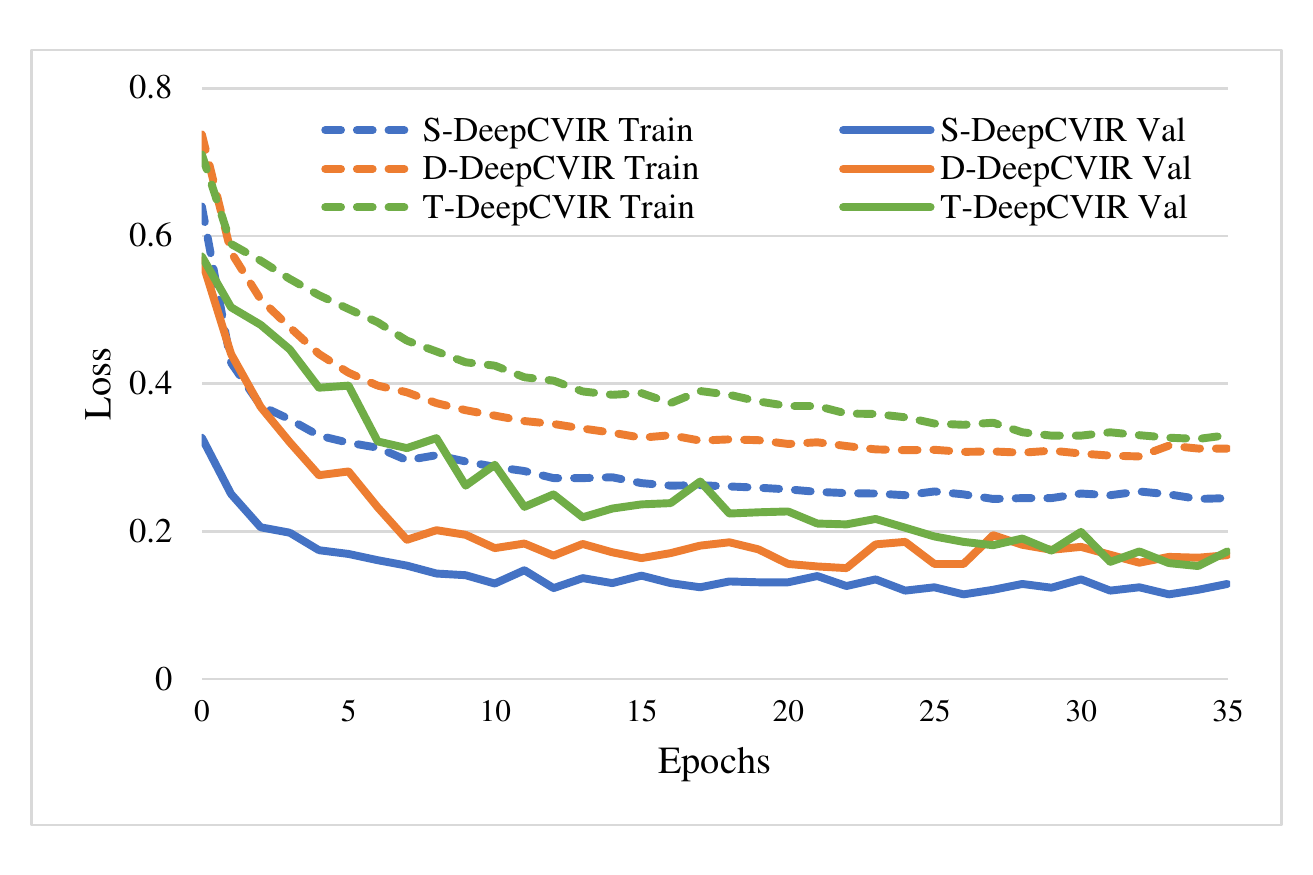}\\
				a) Features vs. Similarity Metric &b) Convergence rate of DeepCVIR
			\end{tabular}
		\end{center}
		\caption{Additional results: a) showing the performance of similarity measuring techniques with various deep supervised features and, b) showing the convergence rate of \{S,T, and D\}-DeepCVIR networks during training and validation. }
		\label{fig:LossPlots}
	\end{figure}

	Various deep features have been previously used for the task of same-view image retrieval; however, view-invariant features of multi-modal images plays a pivotal role in CVIR. Table \ref{Table1:ComparativeFeatures} shows that although Inception-ResNet-v2 may outperform the VGGNet and ResNet on ImageNet challenge yet it failed to extract the most optimal features for cross-view image matching. In addition, the performance of various distance computation methods illustrates that the problem is more complex and could not be solved by linear distances i.e. Euclidean or Contrastive loss embedding. Figure \ref{fig:LossPlots}(a) also confirms the improvement of learning behavior in term of percentage validation accuracy.
		
\begin{table}[b]
	\begin{center}
		\caption{Comparison of different variations of proposed architecture.}\label{Table2:DeepCVIR}
		\resizebox{\linewidth}{!}{%
		\begin{tabular}{|l|c|c|c|c|c|c|c|c|}
			\hline
			DML Type & Feature Type&\multicolumn{1}{c|}{\begin{tabular}[c]{@{}c@{}}Activation \\ Function\end{tabular}} & ANMRR $\downarrow$ & mAP$\uparrow$ & p@5$\uparrow$ & Precision & Recall & F1-Score\\
			\hline\hline
			S-DeepCVIR & \multirow{3}{*}{\centering ResNet-50} & eLU &0.04 &0.93 &0.94 &0.50 &0.50 &0.50\\
			S-DeepCVIR &  & Leaky ReLU &0.03 &0.93 &0.94 &0.94 &0.94 &0.94\\
			
			S-DeepCVIR &  & ReLU &0.03 &0.93 &0.95 &0.93 &0.93 &0.93\\
			\cdashline{1-9}[1.0pt/2pt] 
			\addlinespace[0.1cm]
			D-DeepCVIR& \multirow{2}{*}{\centering ResNet-50} & ReLU &0.04 &0.92 &0.94 &0.92 &0.92 &0.92\\
			T-DeepCVIR&  & ReLU &0.05 &0.89 &0.92 &0.90 &0.90 &0.90\\
			\cdashline{1-9}[1.0pt/2pt] 
			\addlinespace[0.1cm]
			S-DeepCVIR & \multirow{3}{*}{\centering VGG-16} & Leaky ReLU &\textbf{0.02} &\textbf{0.96} &\textbf{0.97} &\textbf{0.96} &\textbf{0.96} &\textbf{0.96}\\
			D-DeepCVIR& & Leaky ReLU &0.02 &0.95 &0.97 &0.95 &0.95 &0.95\\
			T-DeepCVIR& & Leaky ReLU &0.02 &0.96 &0.98 &0.95 &0.95 &0.95\\
			\hline
		\end{tabular}}
	\end{center}
\end{table}

	\subsection{Feature Matching through DeepCVIR}
	The proposed DeepCVIR architecture involves the contribution of DML network which assists the learning process by efficient learning of the embedding space to discriminate similar and dissimilar pairs. However, to evaluate the learning routine of the DML network we tried variants of DML with the single, double and triple combination of the proposed residual blocks termed as S-DeepCVIR, D-DeepCVIR, and T-DeepCVIR, respectively.
	\subsubsection{Impact of the Number of Residual Blocks in DeepCVIR}
	In general increasing the number of residual blocks in a network supports the overall performance; however, in our case S-DeepCVIR with least number of residual blocks outperforms the rest of the DeepCVIR networks. This was beyond our anticipation, but one cannot neglect the simplicity of this task as compare to other recognition tasks. It could be concluded that the number of learnable parameters of S-DeepCVIR are enough to separate similar and dissimilar features. ANMRR and mAP values in Table \ref{Table2:DeepCVIR} illustrates that although all the variants of DeepCVIR performed better than other matching techniques still S-DeepCVIR performed extraordinarily for the given task of Str2Sat as well as Sat2Str. Their convergence curves illustrated in Figure \ref{fig:LossPlots}(b) due to their less number of learnable parameters, represents significantly earlier and much lower loss with respect to the number of epochs as compare to rest of the combinations.
	\subsection{Str2Sat vs. Sat2Str Evaluation}
	Our proposed S-DeepCVIR framework performs equally well on both Str2Sat and Sat2Str tasks. Results in Table \ref{Table3:ComparativeTasks} shows that although the average ANMRR values remain comparative for all the variants of DeepCVIR architecture still S-DeepCVIR with VGG features achieves minimum average ANMRR of $0.025$ and maximum mAP score.
		\begin{figure}[t]
		\begin{center}
			\begin{tabular}{p{.24\textwidth}p{.24\textwidth}p{.24\textwidth}p{.24\textwidth}}
				\includegraphics[scale=0.24]{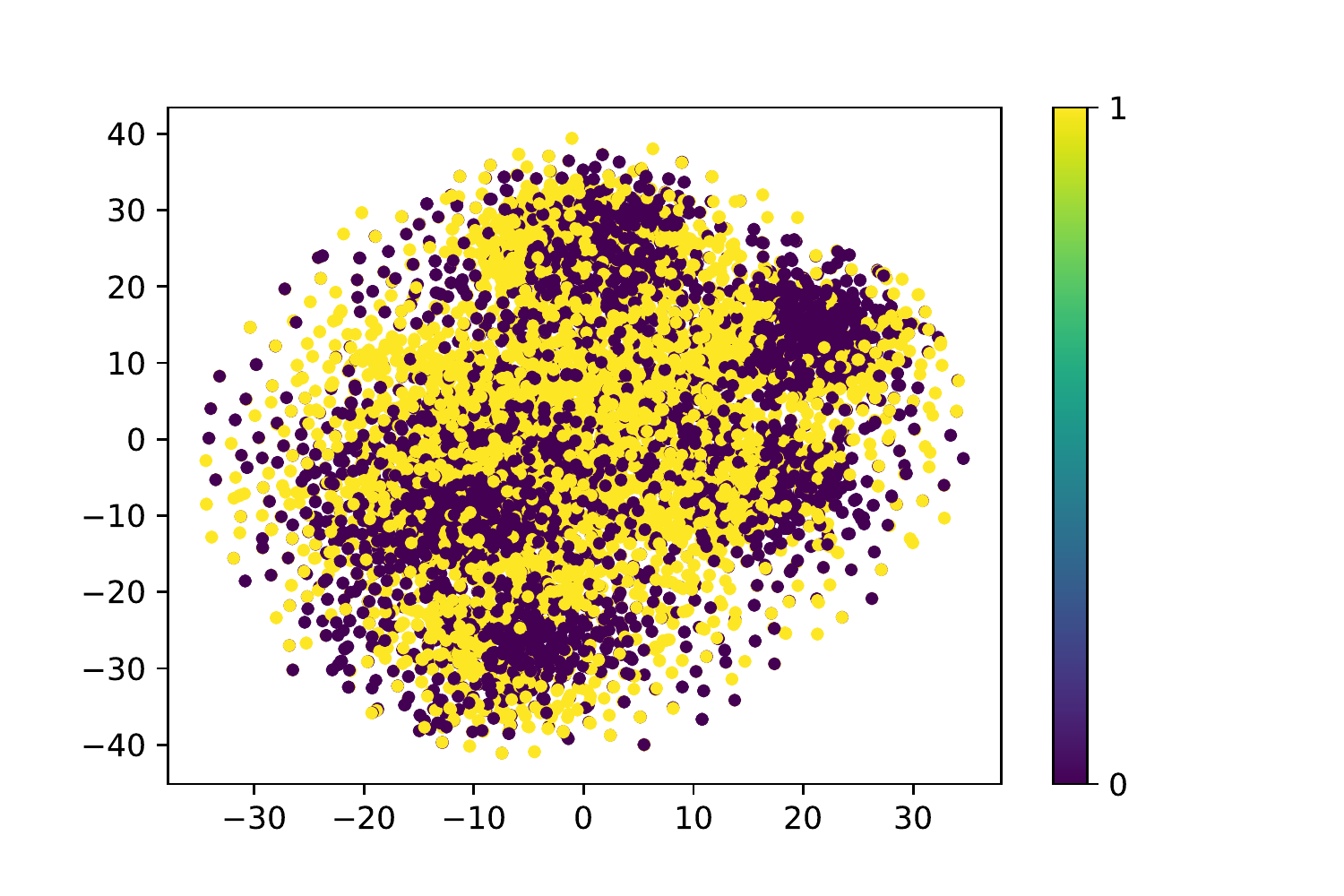}&
				\includegraphics[scale=0.24]{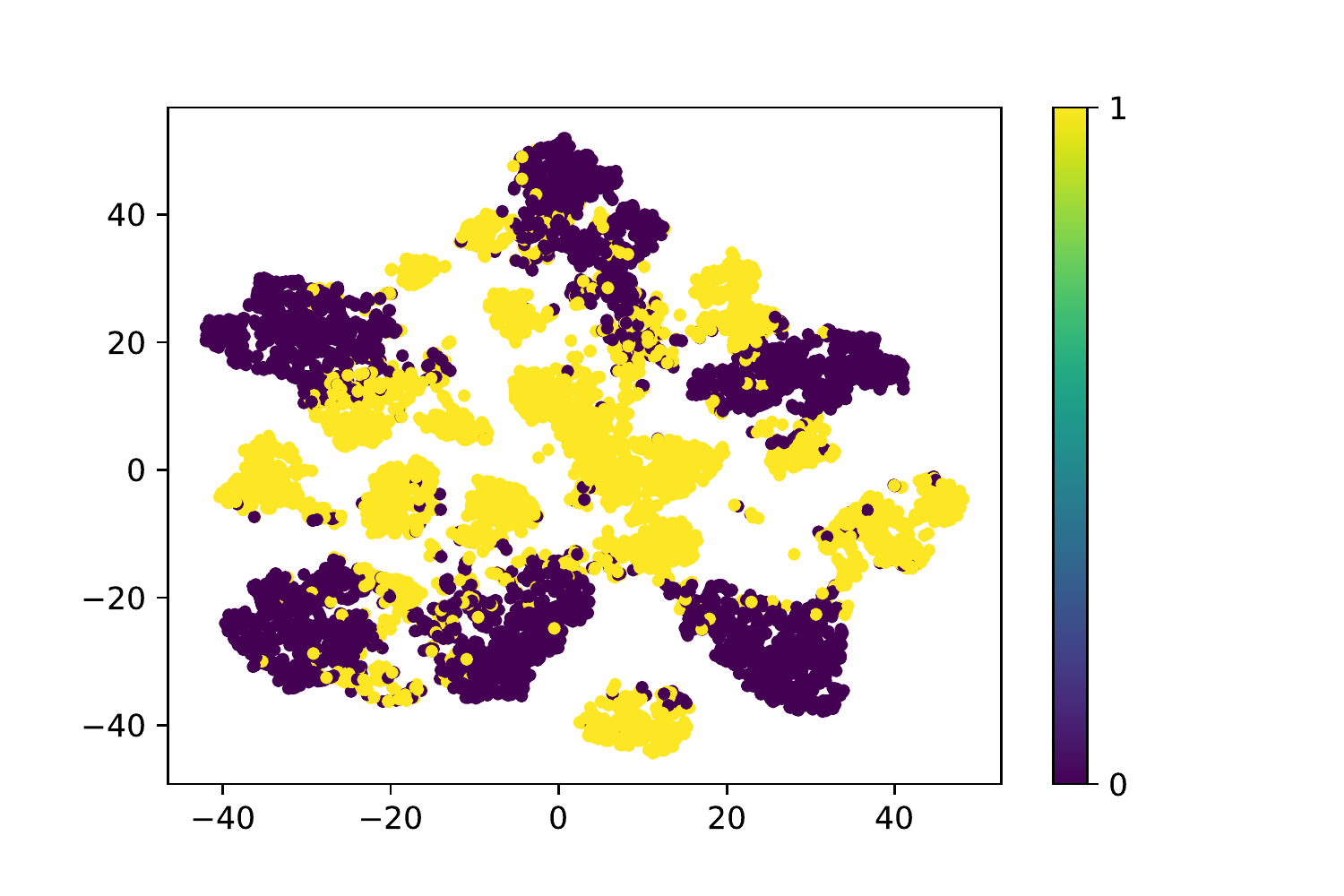}&
				\includegraphics[scale=0.24]{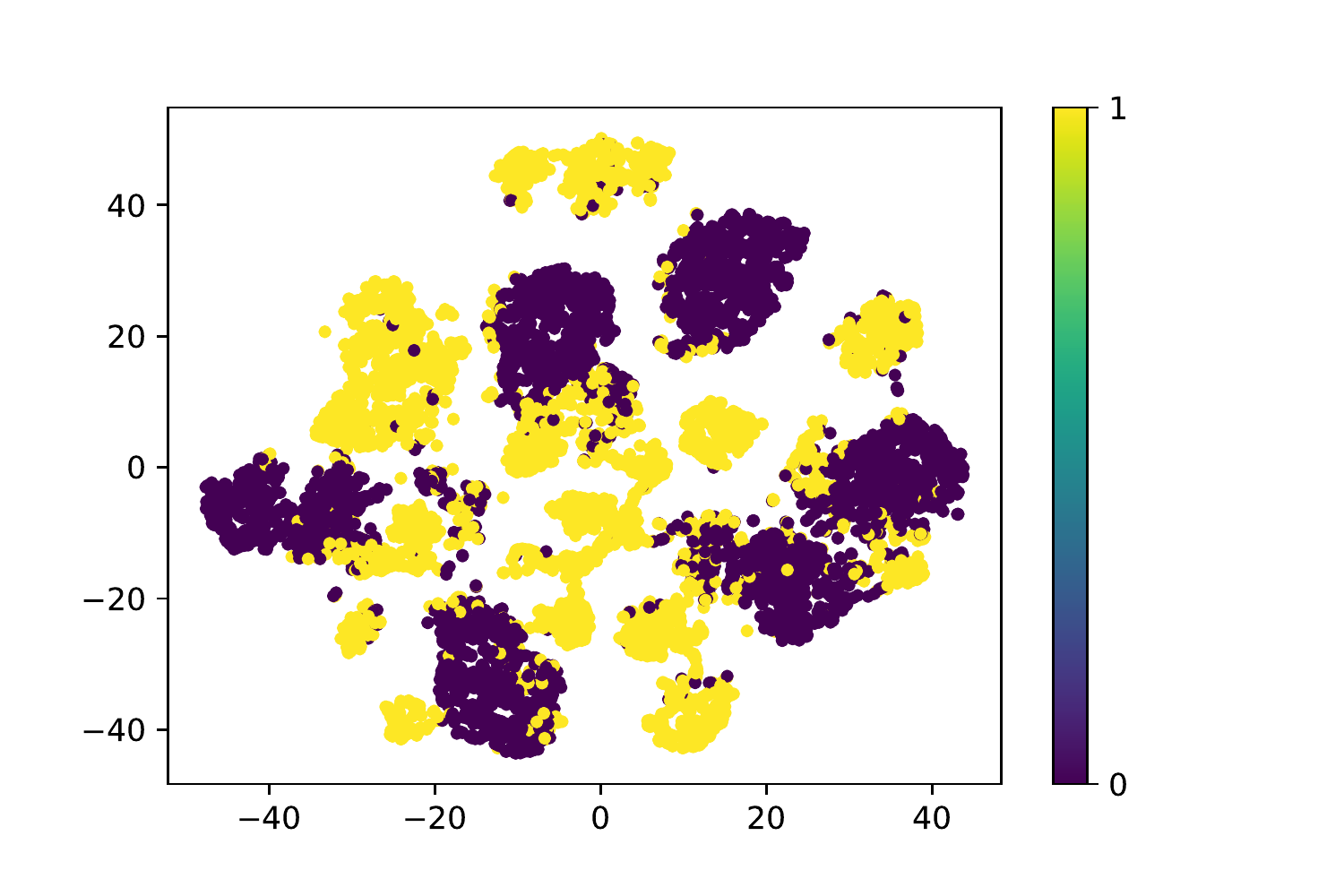}&
				\includegraphics[scale=0.24]{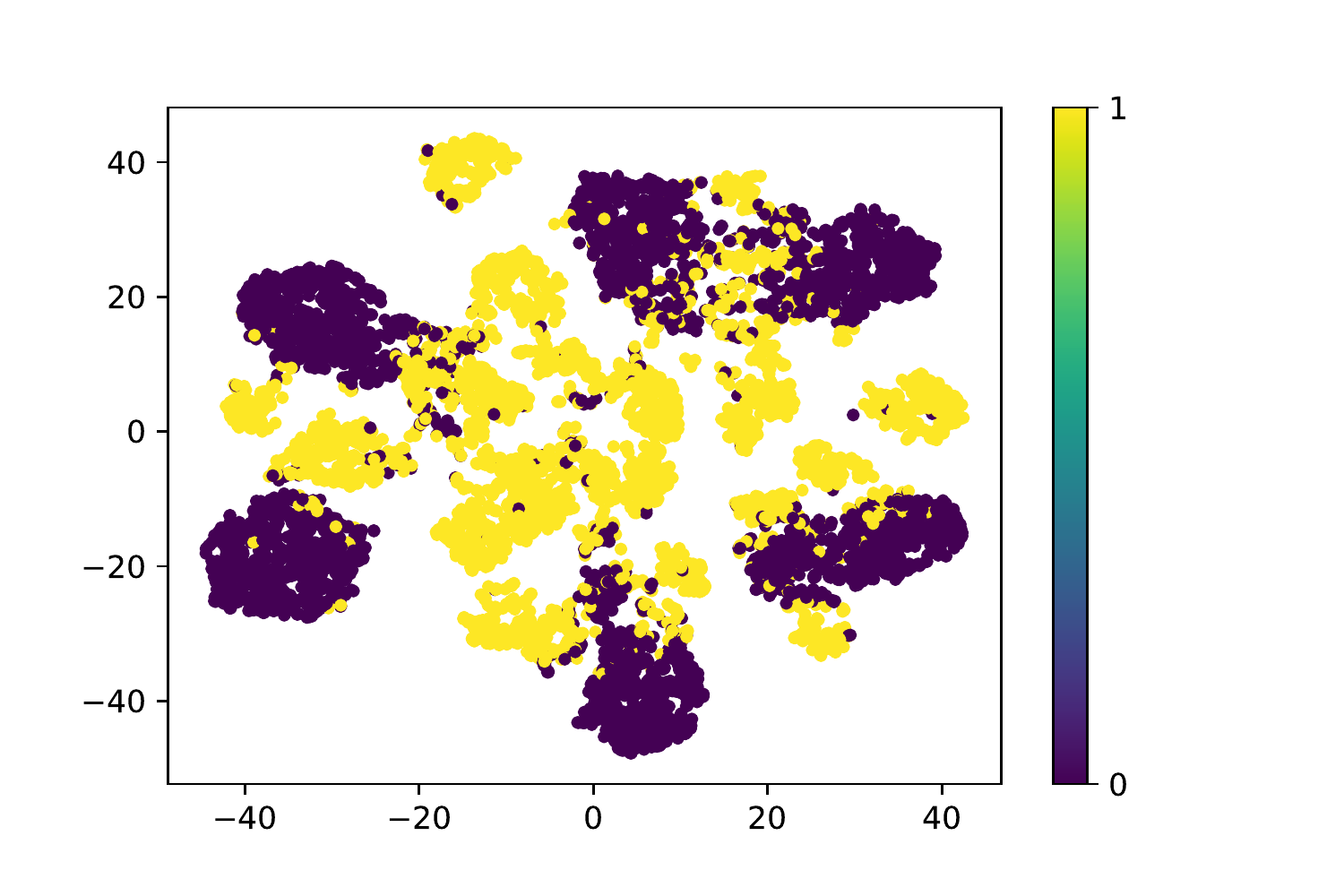}\\
				\hfil a) ResNet-50 &\hfil b) ReLu &\hfil c) Leaky ReLu &\hfil d) eLu\\
				\includegraphics[scale=0.24]{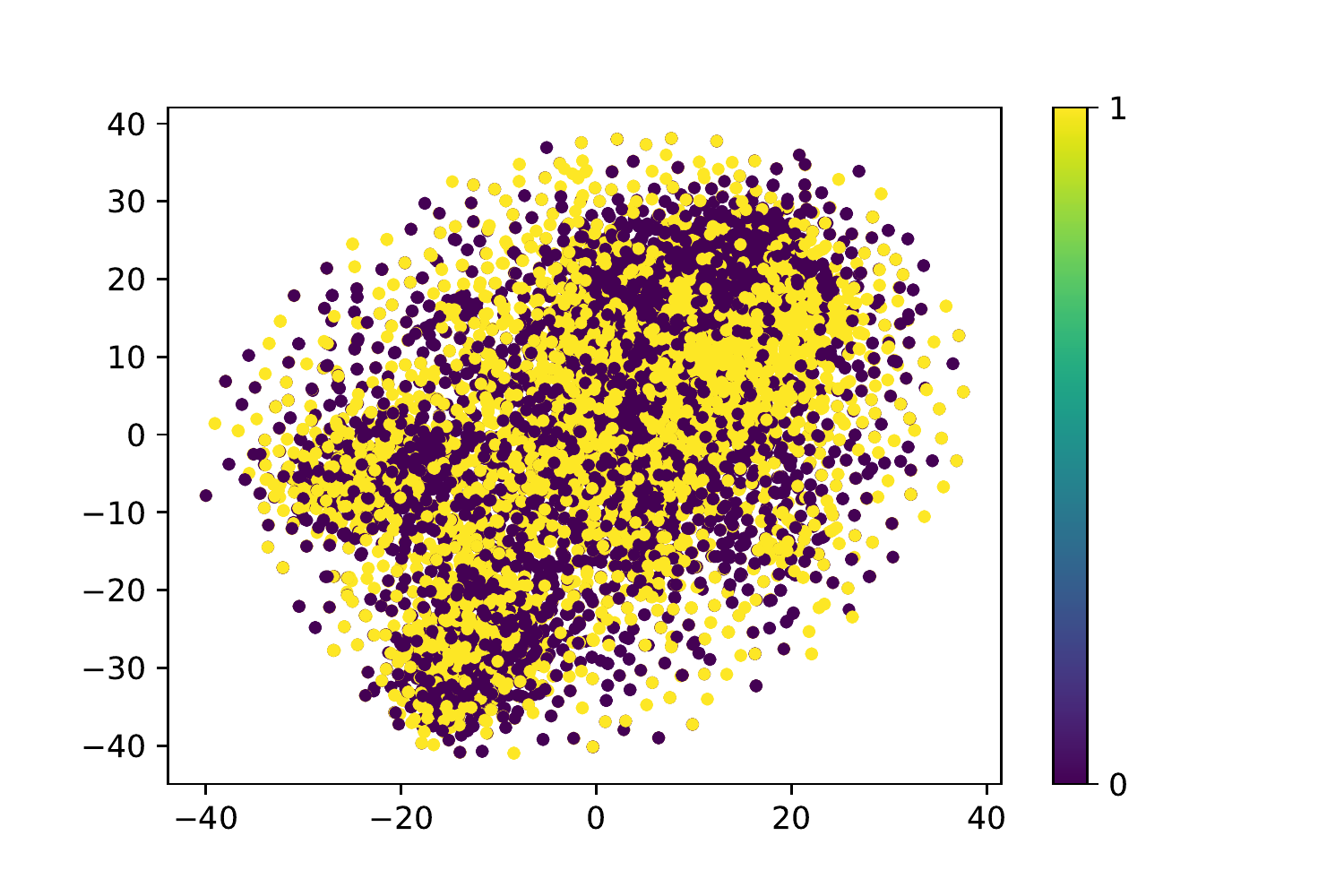}&
				\includegraphics[scale=0.24]{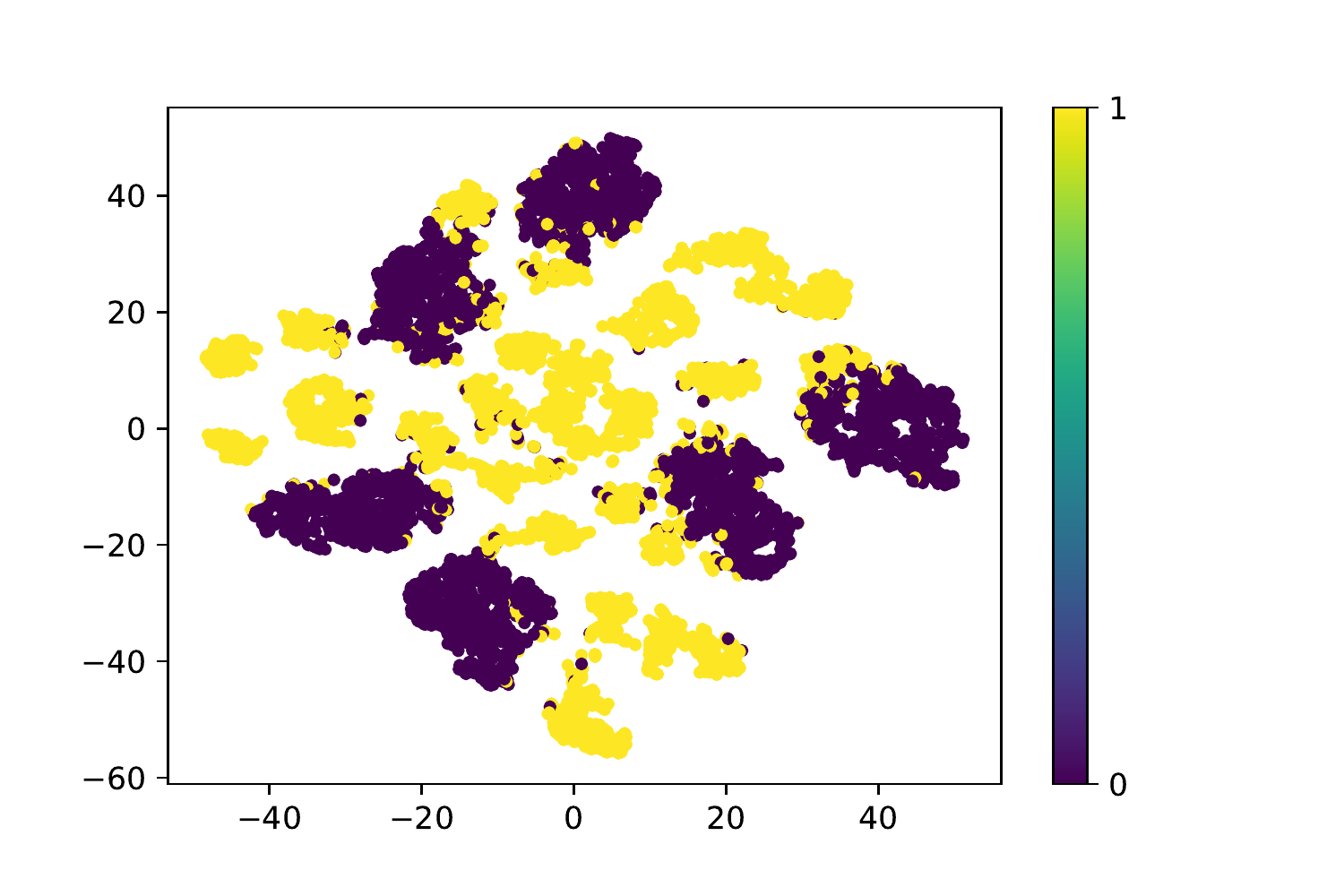}&
				\includegraphics[scale=0.24]{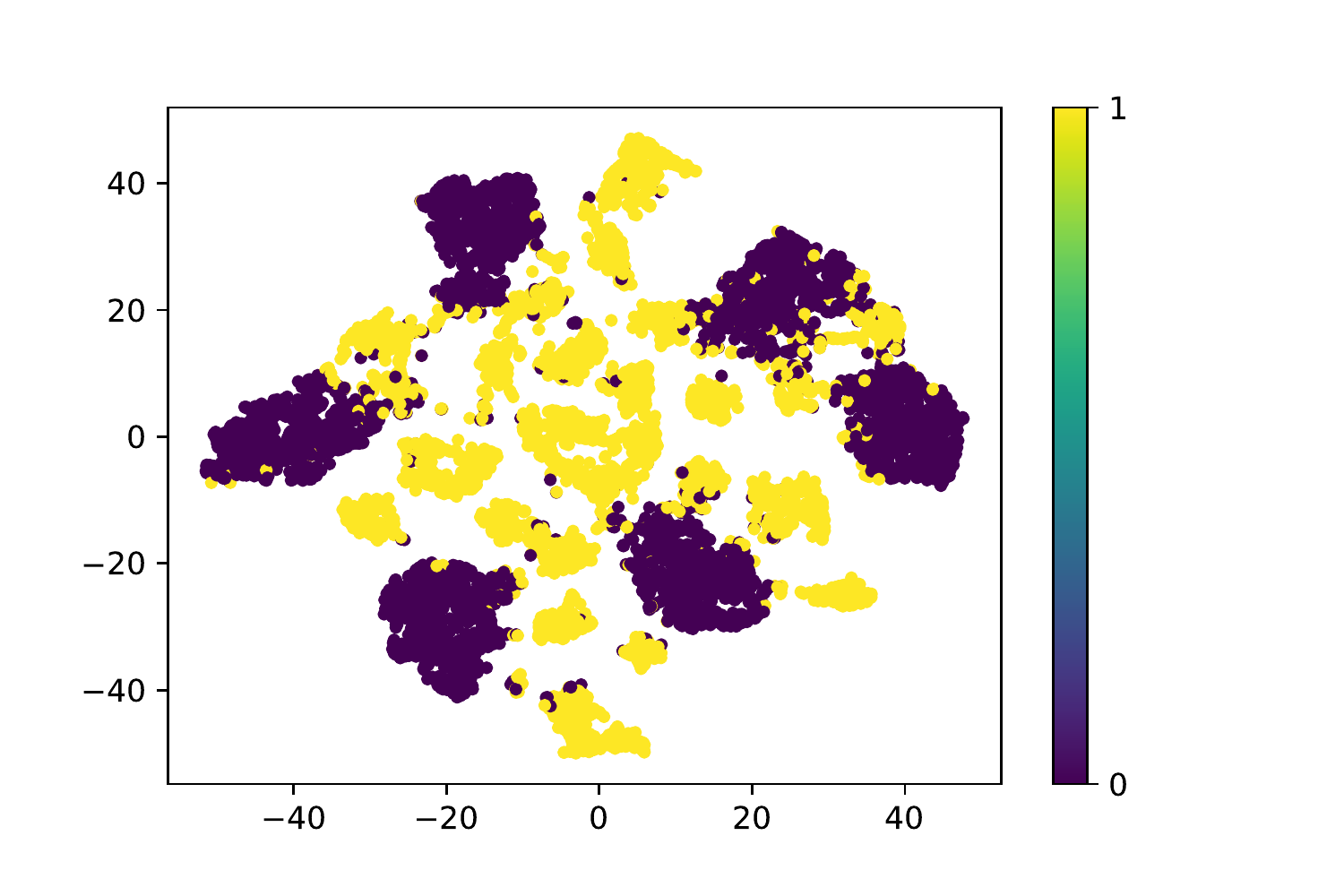}&
				\includegraphics[scale=0.24]{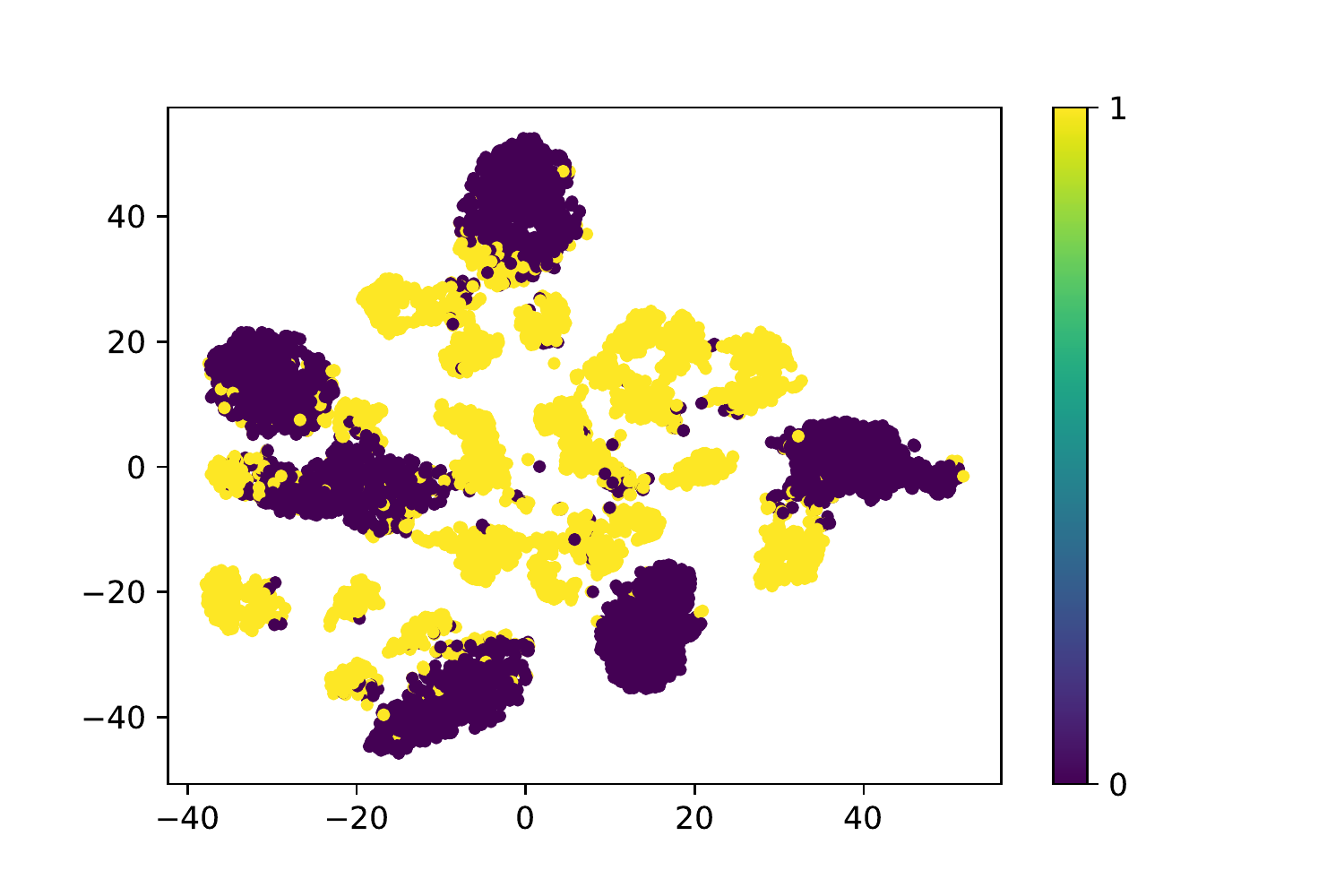}\\
				\hfil e) VGG-16 &\hfil f) S-DeepCVIR &\hfil g) D-DeepCVIR &\hfil h) T-DeepCVIR
			\end{tabular}
		\end{center}
		\caption{t-SNE plots showing learning behavior of DML for: ResNet-50 features trained with S-DeepCVIR employing various activation functions (in top row), and VGG-16 features using different number of residual blocks in DML network (in bottom row).}
		\label{fig:TSNE Plots}
	\end{figure} 
		\begin{table}[b]
	\begin{center}
		\caption{Performance comparison of VGG-16 features with \{S,T, and D\}-DeepCVIR networks for Street-to-satellite (Str2Sat) and Satellite-to-street (Sat2Str) retrieval task.}
		\label{Table3:ComparativeTasks}
		\begin{tabular}{|l|c|c|c|c|c|c|c|c|} 
			\hline
			 \multirow{2}{*}{\centering Features}& \multirow{2}{*}{\centering DeepCVIR}&\multicolumn{3}{c|}{Str2Sat}&\multicolumn{3}{c|}{Sat2Str}&\multirow{2}{*}{\begin{tabular}[c]{@{}c@{}}Average \\ ANMRR\end{tabular}}\\
			 \cmidrule{3-8}
			&  &  ANMRR$\downarrow$ & mAP$\uparrow$ & p@5$\uparrow$& ANMRR$\downarrow$ & mAP$\uparrow$ & p@5$\uparrow$& \\
			\hline\hline
			
			VGG-16 &S-DeepCVIR&\textbf{0.02} &\textbf{0.96} &\textbf{0.97}&\textbf{0.03} &\textbf{0.92} &\textbf{0.92}&0.025\\
			VGG-16 & D-DeepCVIR &0.02 &0.95 &0.97 &0.03 &0.91 &0.91 &0.025\\
			VGG-16 & T-DeepCVIR &0.02 &0.96 &0.98 &0.03 &0.91 &0.91 &0.025\\
			\hline
		\end{tabular}
	\end{center}
\end{table}
	\subsection{t-SNE Visualization of the Learned Embeddings}
	T-SNE plot is a very effective tool to visualize the data in two dimensional plane for better analysis. We adapted this approach to witness and validate the contribution of DML in transforming features to embedding space. Figure \ref{fig:TSNE Plots}(a,f) shows that image features are distributed among the whole region of the plot and hence it is very difficult to measure the correspondence among same and different feature just by using a linear distance. DML separates them into distinguishable clusters. 

	Although no class information was explicitly provided to the network during training still it successfully clustered the similar pairs into six different classes. It is also observed from the figures that use of different activation functions and multiple residual blocks does not contribute to improvement of the overall result.
	
\section{Conclusion}
\label{Conclusion}
We propose a cross-view image retrieval system for which we developed a cross-view dataset named CrossViewRet. The dataset consists of street-view and satellite-view images for 6 distinct classes having 700 images per class. The proposed DeepCVIR system consists of two parts: a) a fine-tuned deep feature network, and b) a deep metric learning network trained on image pairs from CrossViewRet dataset. Given features for two images, the proposed residual DML network decides if the two images belong to the same class. In addition an ablative study and a detailed empirical analysis on different activation functions and number of residual blocks in DML network have also been performed. This shows that our proposed DeepCVIR network performed significantly well for the problem of cross-view retrieval.
	
	\bibliographystyle{splncs04}
	\bibliography{References}
\end{document}